\documentclass{bmvc2k}

\title{\Large Online Domain Adaptation for Multi-Object Tracking}

\addauthor{Adrien Gaidon}{adrien.gaidon@xrce.xerox.com}{1}
\addauthor{Eleonora Vig}{eleonora.vig@xrce.xerox.com}{1}
\addinstitution{
Computer Vision Group\\
Xerox Research Centre Europe\\
Meylan, France
}

\runninghead{Gaidon \& Vig}{Online Domain Adaptation for Multi-Object Tracking}

\usepackage{url}
\usepackage{graphicx}
\usepackage{amssymb}
\usepackage{amsmath}
\usepackage{multirow}
\usepackage{booktabs}
\usepackage[table]{xcolor}

\def\X{{\bf X}}
\def\x{{\bf x}}
\def\Y{{\bf y}}
\def\z{{\bf z}}

\def\W{{\bf W}}
\def\w{{\bf w}}

\def\vel{{\bf v}}
\def\eg{\textit{e.g.,~}}
\def\cf{\textit{cf.~}}
\def\ie{\textit{i.e.~}}
\def\wrt{\textit{w.r.t.~}}
\def\etal{\textit{et~al.~}}

\usepackage{subfigure}
\usepackage[font=small,labelfont=bf,tableposition=top]{caption}
\usepackage{algorithmic}
\usepackage{algorithm,algorithmic}
\usepackage{wrapfig}

\begin{document}

\maketitle

\begin{abstract}
Automatically detecting, labeling, and tracking objects in videos depends first
and foremost on accurate category-level object detectors.
These might, however, not always be available in practice, as acquiring
high-quality large scale labeled training datasets is either too costly or
impractical for all possible real-world application scenarios.
A scalable solution consists in re-using object detectors pre-trained on
generic datasets.
This work is the first to investigate the problem of on-line domain adaptation
of object detectors for causal multi-object tracking (MOT).
We propose to alleviate the dataset bias by adapting detectors from
category to instances, and back: (i) we jointly learn all target models by
adapting them from the pre-trained one, and (ii) we also adapt the
pre-trained model on-line.
We introduce an on-line multi-task learning algorithm to efficiently share
parameters and reduce drift, while gradually improving recall.
Our approach is applicable to any linear object detector, and we evaluate
both cheap ``mini-Fisher Vectors'' and expensive ``off-the-shelf'' ConvNet
features.
We quantitatively measure the benefit of our domain adaptation strategy on
the KITTI tracking benchmark and on a new dataset (PASCAL-to-KITTI) we
introduce to study the domain mismatch problem in MOT.  \end{abstract}

\section{Introduction}

Tracking-by-detection (TBD), the dominant paradigm for object tracking in
monocular video streams, relies on the observation that an accurate
appearance model is enough to reliably track an object in a video.
State-of-the-art Multi-Object Tracking (MOT)
algorithms~\cite{Breitenstein2011, Pirsiavash2011, Milan2014, Geiger2014,
Hall2014, Collins2014}, which aim at automatically detecting and tracking
objects of a known category, rely on the recent progress on object detection.
Most MOT approaches, indeed, consist in finding the best way to associate
detections to form tracks.  They, therefore, directly rely on object detection
performance.
However, a high-quality detector might not always be available in practice. In
particular, acquiring high-quality large scale labeled training datasets
required to train modern detectors is either too costly or impractical for all
possible real-world application scenarios.

\begin{figure}
\center
\vspace*{-1mm}
\includegraphics[width=1.0\linewidth]{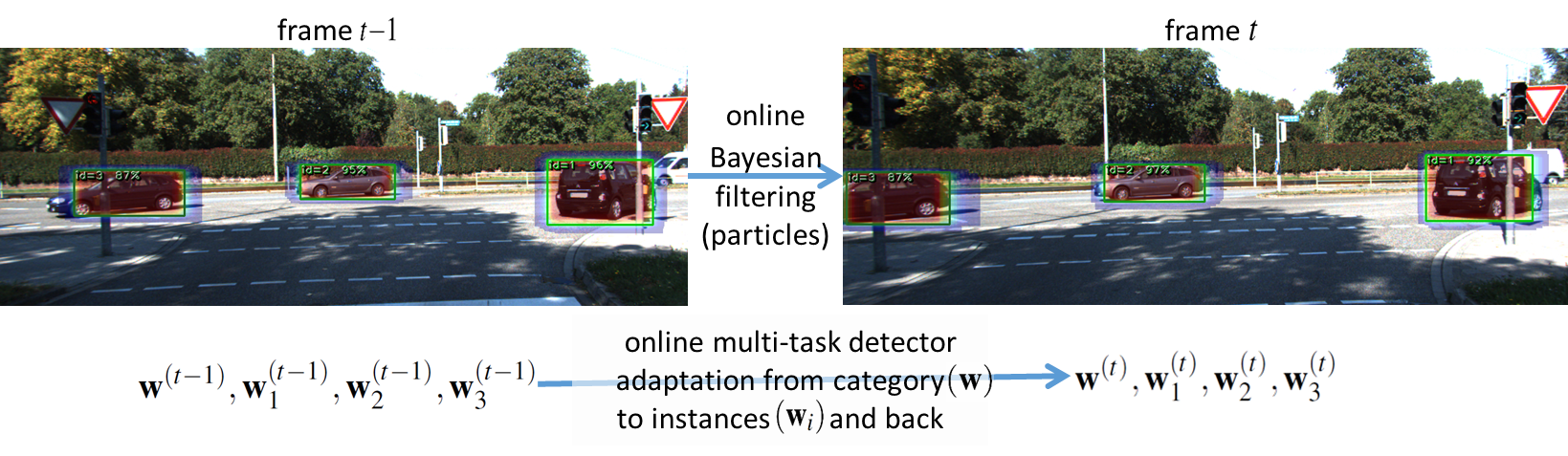} \
\caption{\label{fig:odamot}
Online domain adaptation for MOT via Bayesian filtering coupled with multi-task
adaptation of all detectors jointly.
}
\vspace*{-4mm}
\end{figure}

In this paper, we investigate a scalable solution to this data acquisition
issue: re-using object detectors pre-trained on generic datasets.
We propose to alleviate the ensuing \emph{dataset bias} problem~\cite{bias} for
causal MOT via \emph{on-line domain adaptation of object detectors from
category to instances, and back}.
Previous works (\cf~Section~\ref{s:relwork}) investigated detector adaptation
\emph{or} on-line learning of appearance models, but not both \emph{jointly}.
Our approach can be interpreted as a generalization, where we show that doing
the joint adaptation is key, and doing no adaptation at all significantly
degrades performance because of dataset bias.
We propose a \emph{convex multi-task learning objective} to \emph{jointly
adapt on-line} (i) all trackers from the pre-trained generic detector
(\emph{category-to-instance}), and (ii) the pre-trained category-level
model from the trackers (\emph{instances-to-category}). Our multi-task
formulation enforces parameter sharing between all models to reduce model
drift and robustly handle false alarms, while allowing for continuous
domain adaptation to gradually decrease missed detections.
We integrate our domain adaptation strategy in a novel motion model
combining learned deterministic models with standard Bayesian filtering
(\cf figure~\ref{fig:odamot}) inspired by the popular Bootstrap filter of
Isard \& Blake~\cite{Isard1998}.
In particular, we leverage several techniques not widely used for MOT yet:
(i) recent improvements in object detection based on generic candidate
proposals~\cite{vandeSande2011, Zitnick2014}, (ii) large-displacement
optical flow estimation~\cite{Weinzaepfel2013}, (iii) the Fisher Vector
representation~\cite{perronnin2007fisher, Sanchez2013}, and (iv) ConvNet
features for object detection~\cite{Girshick2014}.
In addition, we use a Sequential Monte Carlo (SMC)
algorithm~\cite{Doucet2000} to approximate the filtering distribution of
our Markovian motion model of the latent target locations.

Section~\ref{s:relwork} reviews the related work. Section~\ref{s:tracklearn}
describes our on-line multi-task learning of the trackers and domain adaptation
of the category-level detector. Our motion model is described in
Section~\ref{s:mot}. Finally, in Section~\ref{s:experiments}, we report
quantitative experimental results on the challenging KITTI tracking
benchmark~\cite{Geiger2012}\footnote{\url{http://www.cvlibs.net/datasets/kitti/eval_tracking.php}}
and on a new \emph{PASCAL-to-KITTI} dataset we introduce for the evaluation of
domain adaptation in MOT.

\section{Related Work}
\label{s:relwork}

Following recent works~\cite{Hall2014, Luo2014, Yang2012},
MOT approaches can be divided into three main categories: (i)
Association-Based Tracking (ABT), (ii) Category-Free Tracking (CFT) and
(iii) Category-to-Instance Tracking (CIT).  

\textbf{ABT} approaches consist in building object tracks by associating
detections precomputed over the whole video sequence. Recent works include
the network flow approach of Pirsiavash~\etal~\cite{Pirsiavash2011}
(DP\_MCF), global energy minimization~\cite{Milan2014} (CEM),
two-granularity tracking~\cite{Fragkiadaki:12}, Hungarian
matching~\cite{Geiger2014}, and
the hybrid stochastic / deterministic approach of Collins and
Carr~\cite{Collins2014}.
These approaches rely heavily on the quality of the pre-trained detector,
as tracks are formed only from pre-determined detections.
Furthermore, they are generally applied off-line and are not always
applicable to the streaming scenario.

\textbf{CFT} approaches, \eg~\cite{Hare2011, Li2013b, Zhang2014, Hua2014},
can be considered as an extension of the category-free single target
approaches to the MOT setting.  In the single target case, the initial
target bounding box is given as input, and a specialized tracker is learned
on-line, \eg via the Track-Learn-Detect approach~\cite{Kalal2011}.
The MOT extension consists in learning different trackers independently for
each target automatically initialized by a generic pre-trained detector,
while also handling the inter-target interactions.
A strength of CFT methods is that they can track any type of object, as
long as their location can be automatically initiated.

\textbf{CIT} approaches are similar to CFT ones in that they learn
independent instance-specific trackers from automatic detections, but the
target-specific models correspond to specializations of the generic
category-level model. This requires the pre-trained detector and the
target-specific trackers to have the same parametric form (\ie same features
and classifier) that work well both at the category and instance levels.
This idea was recently introduced by Hall and Perona~\cite{Hall2014} to
track pedestrians and faces by intersecting detections from a generic
boosted cascade with a target-specific fine-tuned version of the cascade.

Our method is labeled \textbf{ODAMOT}, for ``Online Domain Adaption for
Multi-Object Tracking'' (\cf figure~\ref{fig:odamot}), as it combines
category-to-instance tracker adaptation with a novel (i) multi-task
learning formulation (Section~\ref{s:multitask}) and (ii) algorithm for
on-line domain adaptation of the generic detector
(Section~\ref{s:detadapt}). To the best of our knowledge, our approach is
the first MOT approach to perform domain adaptation of the generic
category-level detector.

Related to our work, Breitenstein~\etal\cite{Breitenstein2011} track
automatically detected pedestrians using a boosted classifier on low-level
features to learn target-specific appearance models.
Another related approach~\cite{Luo2014b} uses a multi-task objective to learn
jointly a generic object model and trackers. It, however, does not use a
pre-trained detector, but initializes targets by hand for each video, assuming
that instances form a crowd of slow-moving near duplicates.
Other related works~\cite{Tang2012, Wang2012, Wang2013} include approaches for
domain adaptation from generic to specific scene detectors for similar
scenarios, although they do not learn trackers.
Some other works~\cite{Sharma2012, Sharma2013, Gaidon2014} do not address
MOT but nonetheless perform detector adaptation specifically for videos via
other means. For instance, \cite{Gaidon2014} puts forth a procedure to
self-learn object detectors for unlabeled video streams by making use of a
similar multi-task learning formulation. On the other hand,
\cite{Sharma2012} relies on unsupervised multiple instance learning to
collect online samples for incremental learning.
Finally, adaptive tracking methods often adopt selective update strategies
to avoid drift, for instance by integrating unlabeled data in the model in
a semi-supervised manner~\cite{Grabner2008}.

\section{Online adaptation from Category to Instances, and back}
\label{s:tracklearn}

\subsection{Generic object detection}
\label{s:god}

\noindent\textbf{Object proposals.}
Current state-of-the-art object detectors (\eg~\cite{vandeSande2011,
Cinbis2013, Girshick2014}) avoid exhaustive sliding window searches.
Instead, they use a limited set of category-agnostic object location
proposals, generated using general properties of objects (\eg contours),
and overlapping most of the objects visible in an image.
Although prevalent in detection, object proposals have not found their way
into multi-object tracking yet. Nevertheless, the advantages of employing
object proposals in MOT are apparent. Since proposals are category- and
target-agnostic, we can reuse feature computations across all detectors
(for any target and category). The speed-up is all the more apparent when
many targets (of possibly different categories) must be tracked.
In addition, object proposals seem well-suited for domain adaptation.
Since object proposal methods rely on generic properties of objects, such
as edge and contour density, they are, indeed, inherently agnostic to the
data source. 
We here adopt the Edge Boxes of Zitnick and Dollar~\cite{Zitnick2014}, as
they yield a good efficiency / accuracy trade-off (\cf~\cite{Hosang2014} for
an extensive review and evaluation of existing proposal methods).
We extract around 4000 object proposals per frame. 

\noindent\textbf{Visual features.}
To represent candidate proposals, we explore the two most common image
representations in current state-of-the-art object detectors with
proposals: Fisher Vectors (FV)~\cite{Cinbis2013} and features from
pre-trained Convolutional Networks~\cite{Girshick2014}. In addition to
being good representations for object detection, they are efficient for
both image classification~\cite{Sanchez2013, Krizhevsky2012} and
retrieval~\cite{jegou2012pami, Babenko2014}. This highlights their
potential for both category-level and instance-level appearance modeling.
Our FV implementation follows~\cite{Cinbis2013}.  We differ, however, by
using only a \emph{single} Gaussian FV (which we call ``mini-FV''), a way
to drastically reduce FV dimensionality (to $2176$ in our case), while
maintaining acceptable performance, as shown for retrieval
by~\cite{Perronnin2010b}.
Regarding the ConvNet features, we follow R-CNN~\cite{Girshick2014}, except
that we replace the standard AlexNet FC7 features with the smaller
$1024$-dimensional features from the penultimate layer of the more
memory-efficient GoogLeNet convolutional network~\cite{Szegedy2015}.
Higher-dimensional representations generally yield higher recognition
performance, but at a prohibitive cost in terms of both speed and memory.
The problem is further exacerbated in MOT, where per-target signatures need
to be persistently stored for re-identification.  We found in our
experiments that the aforementioned features offer a good efficiency and
accuracy trade-off, making them suitable for MOT.
To the best of our knowledge, our method is the first application of FV or
ConvNet features for MOT.

\noindent\textbf{Linear object detector.}
We rank object proposals with a category-specific linear classifier
parameterized by a vector $\w \in \mathbb{R}^d$.  This classifier returns the
probability that a candidate window $\x$, represented by a feature vector
$\phi_t(\x) \in \mathbb{R}^d$, contains an object of the category of interest
in frame $\z_t$ at time $t$ by $P(\x | \z_t; \w) = \left(1 + e^{- \w^T
\phi_t(\x) }\right)^{-1}$.
In our experiments, we estimate the model $\w$ via logistic regression, a
regularized empirical risk minimization algorithm based on the logistic
loss $\ell_t(\x, y, \w) = \log \left( 1 + \exp \left( -y \w^T \phi_t(\x)
\right) \right)$, as this gives calibrated probabilities and has useful
properties for on-line optimization~\cite{Bach2013}.

\subsection{Adaptation from category to instances: multi-task tracking}
\label{s:multitask}

\noindent\textbf{Tracker warm-starting.}
The first category-to-instance adaptation happens at the creation of a new
track. In addition to initializing the target location from a top detection,
in frame $t_0$, we \emph{warm-start} the optimization of the target-specific
appearance model $\w_i^{(t_0)}$ from the category-level one $\w^{(t_0)}$.
Warm-starting allows to start the target optimization close to an
already good solution, as it was used to detect the initial target location.
This yields two positive effects: faster convergence and stronger
regularization. Therefore, warm-starting effectively mitigates the lack of
training data due to the causal nature of our tracker, where we learn models
from a single frame at a time.
Note that warm-starting is often not possible in common MOT approaches, which
generally rely on incompatible features and classifiers (\eg HOG+SVM and
boosted cascades on low-level features~\cite{Breitenstein2011}).

\noindent\textbf{Multi-task regularization.}
Our second adaptation from category to instances consists in \textit{updating
all target models jointly} using multi-task learning. This allows all targets
to share features, reflecting the fact that they belong to the same object
category.
Let $N_t$ be the number of object instances tracked at time $t$.  Each target
$i=1, \cdots, N_t$ has a location $\hat{\x}_i^{(t)}$ predicted by its
associated motion model in frame $t$ (\cf Section~\ref{s:mot}), and a learned
appearance model $\w_i^{(t-1)}$.  The goal is to update this appearance model
$\w_i^{(t-1)} \rightarrow \w_i^{(t)}$ with the new data from frame $t$ by using
the predicted location.
Let $\{\x_{i,k}, k=1, \cdots, n_i \}$ be the $n_i$ training samples of object
$i$ in frame $t$, where $\hat{\x}_i^{(t)}$ is considered as positive, and
negative windows are sampled according to the common ``no teleportation and no
cloning'' assumption on each target individually~\cite{Kalal2011}.
Let $\W^{(t)}=\{\w_1^{(t)},\ldots,\w_{N_t}^{(t)}\}$ be the stacked
target models, and $(\X^{(t)}, \Y^{(t)})$ be the training samples
and labels mined for all targets in frame $t$.
Updating all appearance models jointly amounts to minimizing the following
regularized empirical risk:
\begin{equation}
    \W^{(t)} =
    \arg\min_{\W} L_t(\X^{(t)},\Y^{(t)},\W) + \lambda \Omega_t(\W)
    \label{eq:risk}
\end{equation}
where the loss $L_t$ and multi-task regularization term $\Omega_t$ are defined
as:
\begin{equation}
    L_t(\X^{(t)},\Y^{(t)},\W) =
    \frac{1}{N_t} \sum_{i=1}^{N_t} \frac{1}{n_i} \sum_{k=1}^{n_i}
\ell_t(\x_{i,k}\ , \ y_{i,k}\ , \ \w_i)
	\ , \ \ \ 
    \Omega_t(\W) =
    \frac{1}{2N_t}\sum_{i=1}^{N_t} \|\w_i - \bar{\w}^{(t-1)} \|_2^2,
    \label{eq:reg}
\end{equation}
where $\bar{\w}^{(t-1)}$ is the (running) mean of all previous instance models,
which comprises all past values of the models of currently tracked or now lost
targets (this is equivalent to summing all pairwise comparisons between
target-specific models).
This formulation is closely related to the mean-regularized multi-task learning
formulation of Evgeniou and Pontil~\cite{Evgeniou:04}, with the difference that
it is designed for on-line learning in streaming scenarios.

\subsection{Online adaptation from instances back to category}
\label{s:detadapt}

Our joint multi-task adaptation of the target-specific models allows to track
more reliably while limiting model drift and false alarms thanks to feature
sharing and joint regularization.
In addition, we hypothesize that maintaining and adapting the generic
pre-trained category-level detector should allow to lower the miss-rate by
continuously specializing the global appearance model to the specific video
stream, which might be non-stationary and significantly different than the
off-line pre-training data.
In fact, one can observe that our regularization term (Eq.~\ref{eq:reg})
already provides a theoretical justification to using the running average
$\bar{\w}^{(t)}$ as a single category-level detector.
Indeed, once the detectors $\w_i$ are updated in frame $t$, a new
scene-adapted detector is readily available as:
\begin{equation}
\bar{\w}^{(t)} = 
\frac{1}{\bar{N}_{t-1} + N_t}
\left(
\bar{N}_{t-1} \bar{\w}^{(t-1)}
+
\sum_{i=1}^{N_t} \w_i^{(t)}
\right) \ , \quad
\text{where} \ \ \bar{N}_{t-1} = \sum_{j = 1}^{t-1} N_j.
\label{eq:final}
\end{equation}

As we use linear classifiers, this multi-task learning is akin to a ``fusion''
of exemplar-based models (\eg Exemplar-SVMs~\cite{exemplar}). A major
improvement is that our models are learned \emph{jointly} and \emph{adapt
continuously} to both the data stream and other exemplars.
This adaptation allows to limit drift of the category model. There is,
indeed, an ``inertia'' in the update due to the warm-starting of the
trackers from the generic model. Furthermore, as the adapted model
corresponds to a (potentially long) running average, the contribution of
false alarms to the model should be limited, as false alarms are more
likely to be tracked for less time thanks to our multi-task penalization.
We optimize the learning objective in Eq.~\ref{eq:risk} using Stochastic
Gradient Descent (SGD) with constant learning rate of $10^{-5}$.

\section{Causal Multi-Object Tracking-by-Detection}
\label{s:mot}

In this section, we describe our causal (\ie on-line) MOT framework to
track a variable number of objects belonging to a category known in advance
(\eg cars) in a monocular video stream coming from a fixed or moving
camera.
Algorithm~\ref{fig:pseudocode} provides a high-level pseudo-code
description of ODAMOT.

\subsection{Bayesian motion model}
\label{s:tbd}

Let $\z_t$ be the random variable representing our observation, a frame of
the video stream at time $t$.
Let $\x_t = (x_t, y_t, w_t, h_t)^T$ be the random variable representing the
latent location (a bounding box) of the object $i$ in frame $\z_t$.
We model the evolution of the object's location using a dynamical system
specific to target $i$ characterized by the following Bayesian model.

\begin{wrapfigure}{R}{0.5\textwidth}
\begin{minipage}{0.5\textwidth}
\vspace{-4mm}
\begin{algorithm}[H]
\small
\begin{algorithmic}
    \STATE \textbf{Input:}
        generic detector $\w$, video stream
    \STATE \textbf{Output:}
        adapted detector $\bar{\w}^{(t_{\mathrm{end}})}$, tracks list
$\mathcal{W}$
    \STATE \textbf{Initialization:} $\bar{\w}^{(t_0)} = \w$, $\mathcal{W} =
           \emptyset$ 
    \WHILE{video stream is not finished}
    \FOR{each target $i$ in $\mathcal{W}$}
        \STATE Update $i$'s location 
with a Bayesian motion model (\cf Sec.~\ref{s:tbd})
    \ENDFOR
    \STATE Detect new targets not in $\mathcal{W}$ with $\bar{\w}^{(t-1)}$
in frame $t$ and add them to $\mathcal{W}$ 
    \STATE Merge overlapping tracks in $\mathcal{W}$
    \FOR{each target $i$ in $\mathcal{W}$}
        \IF{$i$ is a new target}
        \STATE Learn initial detector $\w_i^{(t-1)}$ warm-started from
           $\bar{\w}^{(t-1)}$ (\cf Sec.~\ref{s:multitask}) 
        \ENDIF
        \STATE Run the detector $\w_i^{(t-1)}$
        \IF{object $i$ is lost}
            \STATE Remove $i$ from $\mathcal{W}$
        \ELSE
            \STATE Get $\{(\x_{i,k}^{(t)}, y_{i,k}^{(t)}), \ k=1:n_i \}$
            \STATE Update detector $\w_i^{(t)}$ (\cf
            Sec.~\ref{s:multitask})
        \ENDIF
    \ENDFOR
    \STATE Update generic detector $\bar{\w}^{(t)}$ (Eq.~\ref{eq:final})
    \ENDWHILE
\end{algorithmic}
\caption{Pseudo-code overview of our approach. Refer to the main text for 
details.}\label{fig:pseudocode} 
\end{algorithm}
\vspace{-5mm}
\end{minipage}
\end{wrapfigure}

\noindent The initial distribution is $\x_{t_0} \sim \mathcal{N}(\hat{\x}_{t_0},
\Sigma_{t_0})$, where $\hat{\x}_{t_0}$ is the target's initial location in a
frame $t_0 < t$, which corresponds to a detection of the generic detector
$\w^{(t_0)}$ in frame $t_0$, and $\Sigma_{t_0}$ is the initial covariance
modeling the uncertainty on this initial location.

Our Markovian transition model is $\x_t = \x_{t-1} + \vel_{t-1}(\x_{t-1})
\Delta t + \epsilon_t$, where $\epsilon_t$ is Gaussian noise and
$\vel_{t-1}(\x_{t-1})$ is the instantaneous target velocity estimated by median
filtering in the dense large-displacement optical flow field computed using the
deep flow algorithm~\cite{Weinzaepfel2013}. Note that this differs from the
standard constant velocity assumption~\cite{Breitenstein2011}, which is not
suitable for fast moving objects and moving cameras.

Our observation model is characterized by:
$P(\z_t | \x_t) \propto
P( \x_t | \z_t; \w_i^{(t-1)} )
\cdot
P( \x_t | \z_t; \w^{(t-1)} )$.
We define the likelihood to be proportional to the probability that the window
has both the appearance of target $i$, modeled by the target-specific
appearance model $\w_i^{(t-1)}$, and of the category, modeled by the
category-level appearance model $\w^{(t-1)}$, assuming uniform priors over the
frames $\z_t$ and locations $\x_t$.
The appearance models $\w_i^{(t-1)}$ and $\w^{(t-1)}$ are obtained at the
previous time step $t-1$, as described in the previous
Section~\ref{s:tracklearn}.

\vspace*{-2mm}
\subsection{Sequential Monte Carlo approximation of the filtering distribution}

In order to use this model, we need to recursively estimate the filtering
distribution $P(\x_t | \z_{1:t})$.
Following the standard practice, we approximate the filtering distribution
using Sequential Monte Carlo sampling.
We use Sequential Importance Sampling~\cite{Doucet2000} to compute our
filtering distribution approximation recursively over time using $N$
particles $\x_{t-1}^{(p)}$, $p=1, \cdots, N$.
In practice, we found that $N=100$ particles provided a good trade-off
between exploration, exploitation, and computational efficiency. We use
$\sigma_0 = 5\%$ as fixed initial relative noise variance, and scale it by
the inverse of the number of successful updates for the target.

To predict precisely the location of an object at each time instant from our
estimate of the filtering distribution, we use the expectation of the latent
variable $\x_t$, as it can be easily estimated as the weighted average of the
particles: $\hat{\x}_t = \sum_{p=1}^N w_t^{(p)} \x_t^{(p)}$.
We observed that using the expectation yields good results, as the distribution
tends to have a limited variance due to the specialization of the per-target
appearance models.

\vspace*{-2mm}
\subsection{Inter-target reasoning}
\label{s:mtt}

Our \emph{identification} strategy to MOT differs from standard global data
association methods, as it relies on the detector(s) to limit ID switches and
fragmentation. We also rely mostly on appearance to handle occlusions, as this
sort of invariance is a goal of object detection.
However, as our detectors might suffer from dataset bias, we apply further
post-processing to deal with occlusions.
In particular, we temporarily lose a target, \ie make no location prediction,
and try to reinitialize its location in subsequent frames using its
specialized detector.
If the reinitialization fails consecutively for more than $T$ frames, we
terminate the target. Note that later re-identification can be performed by
trying to reinitialize at bigger regular time intervals.
We also assume that two overlapping tracks correspond to the same target if the
location predictions intersect by more than $30\%$ for more than $T$
consecutive frames. In this case, the tracker with the lower score is
terminated.
In our experiments, we used the very short $T=3$ interval in order to deal
with potentially fast moving objects (cars) filmed from a fast moving
platform (a car-mounted camera).
Note that our main contribution (online joint domain adaptation of all
appearance models) is orthogonal to the numerous occlusion reasoning and data
association improvements to MOT (\eg~\cite{Yoon2015}), which could be combined
with our method for improved performance.

\vspace*{-2mm}
\section{Experiments}
\label{s:experiments}

We evaluate our MOT algorithm on the challenging KITTI car tracking
benchmark~\cite{Geiger2012}. As this challenge discourages multiple
submissions on its evaluation server, we evaluate only the best detector we
can train on related training data using state-of-the-art ConvNet features.
We then perform an ablative analysis and quantitatively demonstrate the
benefit of our domain adaptation strategy on the new PASCAL-to-KITTI (P2K)
dataset, which we describe below.
In our experiments, we follow the KITTI evaluation protocol by using the
CLEAR MOT metrics~\cite{Bernardin2008} and
code\footnote{\url{http://kitti.is.tue.mpg.de/kitti/devkit_tracking.zip}}
-- including MOT Accuracy (MOTA), MOT Precision (MOTP), Fragmentation
(FRG), and ID Switches (IDS) -- complemented by the Mostly Tracked (MT) and
Mostly Lost (ML) ratios, Precision (Prec), Recall (Rec), and False Alarm
Rate (FAR).

\begin{table*}
    \footnotesize 
    \centering
    \setlength{\tabcolsep}{5pt}
    \begin{tabular}{rrrrrrrrrrrrr}
    \rowcolor{gray!50}
    method   &  MOTA$\uparrow$  &  MOTP$\uparrow$  &   MT$\uparrow$  & ML$\downarrow$  &   Rec.$\uparrow$    &  Prec.$\uparrow$    &  FAR$\downarrow$  & IDS$\downarrow$ & FRG$\downarrow$ \\
    \toprule                                                              
DP\_MCF$\dagger$~\cite{Pirsiavash2011} & 36.62\% & 78.49\% & 11.13\% & 39.18\% & 46.19\% & 96.64\% & 5.03\% & 2738 & 3240 \\
HM~\cite{kuhn1955hungarian} & 42.22\% & 78.42\% & 7.77\% & 41.92\% & 43.83\% & 96.80\% & 4.54\% & \textbf{12} & 577 \\
MCF$\dagger$~\cite{Zhang2008} & 44.28\% & 78.32\% & 10.98\% & 39.94\% & 45.87\% & \textbf{97.03}\% & \textbf{4.40}\% & 23 & 590 \\
TBD$\dagger$~\cite{Geiger2014} & 52.44\% & 78.47\% & 13.87\% & 34.30\% & 55.28\% & 95.51\% & 8.16\% & 33 & 538 \\
DCO$\dagger$~\cite{Andriyenko2012} & 35.17\% & 74.50\% & 10.67\% & 33.69\% & 50.74\% & 77.56\% & 46.13\% & 223 & 622 \\
CEM$\dagger$~\cite{Milan2014} & 48.23\% & 77.26\% & 14.48\% & 33.99\% & 54.52\% & 90.47\% & 18.04\% & 125 & 398 \\
RMOT~\cite{Yoon2015} & 49.87\% & 75.33\% & 15.24\% & 33.54\% & 56.39\% & 90.16\% & 19.35\% & 51 & \textbf{385} \\
DCO\_X*$\dagger$~\cite{Milan2013} & \textbf{62.76}\% & \textbf{78.96}\% & 26.22\% & 15.40\% & 77.08\% & 86.47\% & 39.17\% & 326 & 984 \\
RMOT*~\cite{Yoon2015} & 60.46\% & 75.57\% & \textbf{26.98}\% & \textbf{11.13}\% & \textbf{79.19}\% & 82.68\% & 54.02\% &  216 & 742 \\
\rowcolor{gray!25}                                                    
\textbf{ODAMOT} & 57.06\% & 75.45\% & 16.77\% & 18.75\% & 64.76\% & 92.04\% & 17.93\% & 404 & 1304 \\
    \bottomrule
    \end{tabular}
    \caption{KITTI Car tracking benchmark results. Metrics with $\uparrow$
(resp. $\downarrow$ ) should be increasing (resp.\ decreasing).  Methods
with * use Regionlets~\cite{Wang2013regionlets}.  Those with $\dagger$ are
offline, the others online.}
    \label{tab:resKittiTest}
    \vspace*{-3mm}
\end{table*}

\vspace*{-2mm}
\subsection{KITTI tracking benchmark}
\label{s:kitti}
The KITTI object tracking
benchmark~\cite{Geiger2012}\footnote{\url{http://www.cvlibs.net/datasets/kitti/eval_tracking.php}}
consists of 21 training and 29 test videos recor-ded using cameras mounted
on a moving vehicle. This is a challenging dataset designed to investigate
how computer vision algorithms perform on real-world data typically found
in robotics and autonomous driving applications.
We train an R-CNN-like car detector on the 21 training videos for which
ground truth tracks are available (\cf Section~\ref{s:god} for more details).
As in~\cite{Girshick2014}, for increased performance, we perform
domain-specific fine-tuning of the network on the KITTI training set prior
to training the detector.  

\noindent\textbf{Results.}
Table~\ref{tab:resKittiTest} summarizes the tracking accuracy of our method
(\textbf{ODAMOT}) and other state-of-the-art approaches on the 29 test
sequences whose ground truth annotations are not public. We compare against
all the results on this benchmark where the methodology has been described
in the literature.
Our algorithm ranks third in terms of MOTA, which summarizes multiple
aspects of tracking performance. An explanation for the performance gap
lies in the adoption of more sophisticated inter-target and occlusion
reasoning by competing methods~\cite{Milan2013,Yoon2015}.
RMOT~\cite{Yoon2015}, for instance, performs data association and leverages
motion context in addition to Bayesian filtering.  Indeed, the rather
simple inter-target reasoning of ODAMOT explains the high number of ID
switches and fragmentations, which are detrimental to performance.

\subsection{PASCAL-to-KITTI: domain adaptation in MOT}
\label{s:ptk}

\begin{figure}
\center
\vspace*{2mm}
\includegraphics[height=1.64cm]{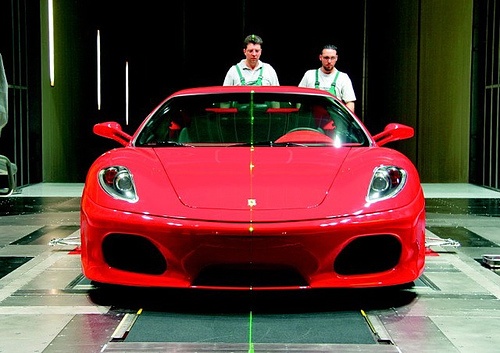} \
\includegraphics[height=1.64cm]{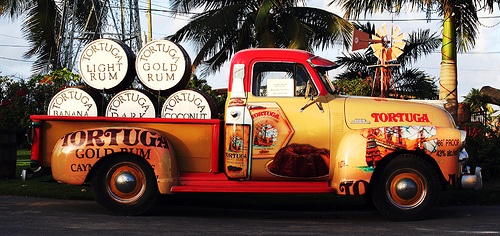} \
\includegraphics[height=1.64cm]{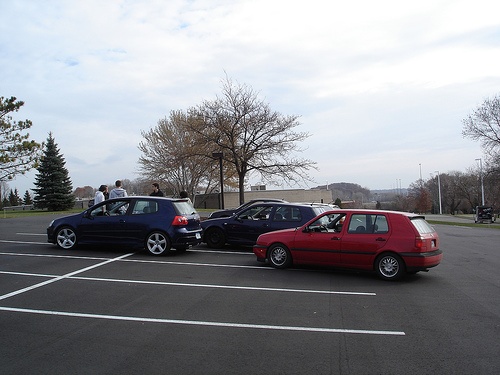} \
\hspace*{1.7mm}\includegraphics[height=2.516cm]{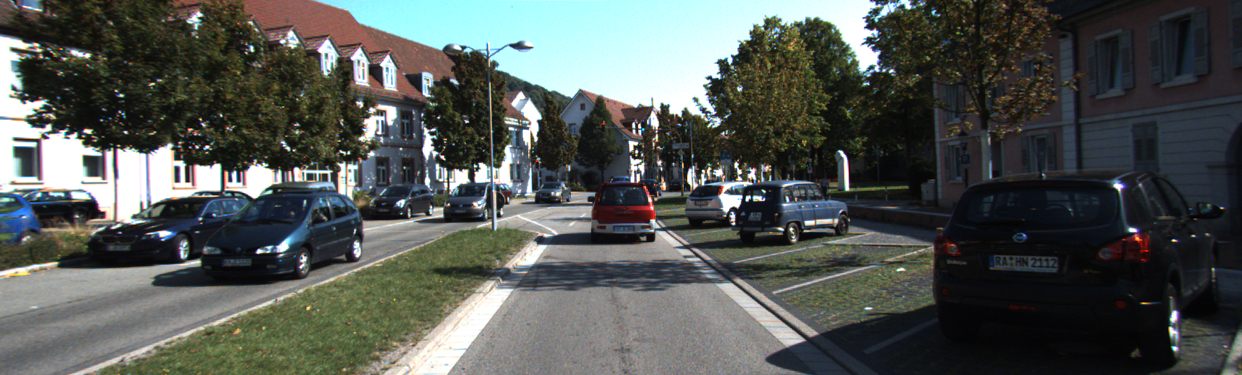} \
\vspace*{2mm}
\caption{\label{fig:data} Images from the Pascal VOC 2007 (top) and KITTI
Tracking (bottom) benchmarks. Note the striking differences in visual
appearance between the two datasets.}
\vspace*{-3mm}
\end{figure}

\noindent\textbf{PASCAL-to-KITTI (P2K) dataset.}
Domain adaptation of appearance models for MOT has remained largely
unaddressed until now. To allow the systematic study of this problem, we
assembled a new MOT dataset called \emph{PASCAL-to-KITTI} (P2K).
The training set (the \emph{source domain}) consists of the training images
of the standard Pascal VOC 2007 detection challenge~\cite{Everingham2010}.
As this dataset is general-purpose, it is reasonable to expect it to yield
pre-trained appearance models that are likely to transfer to more specific
tasks or domains, at least to a certain extent.
The test set (the \emph{target domain}) consists of the 21 training videos
of the KITTI tracking challenge. Fig.~\ref{fig:data} highlights some
striking differences between source and target domains and illustrates the
difficulty of transfer.

\noindent\textbf{Detector pre-training.}
The pre-training of the detector is performed off-line via batch logistic
regression (using liblinear~\cite{liblinear}) with hard negative mining as
described in~\cite{Cinbis2013}.
Our mini-FV model yields $40\%$ Average Precision (AP) for car
detection on the Pascal test set, which is $18\%$ below the results
of~\cite{Cinbis2013} for a fraction of the cost. Our R-CNN-like detector
achieves $60\%$ AP on the Pascal test images, which is on par with the
results reported by~\cite{Girshick2014} ($58,9\%$ AP for R-CNN fc$_7$).
On three validation videos of the KITTI training set this detector gives
$42\%$ AP, which hints at the domain gap between Pascal and KITTI.

\begin{table*}
    \footnotesize
    \centering
    \begin{tabular}{rrlrrrrrrrrr}
    \rowcolor{gray!50}
    method   & MOTA$\uparrow$  &  MOTP$\uparrow$  &   MT$\uparrow$  & ML$\downarrow$  &   Rec.$\uparrow$    &  Prec.$\uparrow$    &  FAR$\downarrow$  & IDS$\downarrow$ & FRG$\downarrow$ \\
    \toprule                                                              
    DP\_MCF$\dagger$~\cite{Pirsiavash2011}                                        
             &  1.9\% & 74.0\% & 0.0\% & 98.6\% &  2.1\% & \textbf{92.9\%} & \textbf{0.5\%} &   \textbf{6} &   \textbf{25} \\
    G\_TBD$\dagger$~\cite{Geiger2014}
             &  8.4\% & 71.2\% & 0.2\% & 86.9\% & 11.1\% & 81.0\% &  8.1\% & 13 & 174 \\
       CFT~\cite{Kalal2011}  & 16.6\% & \textbf{74.9\%} & \textbf{1.1\%} & 71.8\% & 19.2\% & 88.0\% &  8.1\% & 68 &  254 \\ 
       CIT~\cite{Hall2014}    & 18.2\% & 73.9\% & \textbf{1.1\%} & 67.3\% & 21.8\% & 86.1\% & 10.9\% &  40 &  193 \\  
    \rowcolor{gray!25}                                                    
    \textbf{ODAMOT}                                                       
             & \textbf{19.7\%} & 74.5\% & \textbf{1.1\%} & \textbf{64.6\%} & \textbf{23.5\%} & 86.4\% & 11.5\% &  55 &  232 \\ 
    \midrule
    DP\_MCF$\dagger$~\cite{Pirsiavash2011}                                         
           &  12.0\% & 68.5\% & 0.1\% & 80.2\% & 14.6\% & \textbf{85.5}\% &  \textbf{7.7}\% &  \textbf{84} &  \textbf{327} \\
    G\_TBD$\dagger$~\cite{Geiger2014}
           &  17.5\% & 68.0\% & 0.9\% & 59.2\% & 30.0\% & 71.3\% & 37.6\% & 115 &  528 \\
       CFT~\cite{Kalal2011} & 17.6\% & 66.7\% & 1.8\% & 45.7\% & 33.5\% & 69.1\% & 47.2\% & 238 &  592 \\ 
       CIT~\cite{Hall2014}  & 22.8\% & 68.5\% & \textbf{1.9}\% & \textbf{43.4}\% & 33.9\% & 76.5\% & 32.6\% & 380 &  809 \\ 
    \rowcolor{gray!25}                                                    
    \textbf{ODAMOT}                                                       
             & \textbf{23.6}\% & \textbf{68.7\%} & 1.8\% & 43.6\% & \textbf{34.2}\% & 77.5\% & 31.1\% & 376 &  784 \\ 
    \bottomrule
    \end{tabular}
    \caption{MOT results on the P2K domain adaptation dataset.
        The upper block contains results for the ``mini-Fisher Vector'' detector,
        while the lower block shows results for the more powerful R-CNN-like
        detector. Methods with $\dagger$ are offline, the others are online.}
    \label{tab:rescar}
    \vspace*{-3mm}
\end{table*}

\noindent\textbf{Baselines.}
We compare \textbf{ODAMOT} to the related MOT algorithms from
Section~\ref{s:relwork}: off-line Association Based Tracking (ABT) type
methods (DP\_MCF~\cite{Pirsiavash2011}, and G\_TBD~\cite{Geiger2014},
for which code is available), and our implementation of an on-line
Category-Free Tracker (CFT) and an on-line Category-to-Instance Tracker
(CIT). CFT corresponds to the TLD approach of~\cite{Kalal2011}, and differs
from ODAMOT in that it does not warm-start the target models from a
pre-trained detector, performs no multi-task regularization (target models
are independent), and no online adaptation of the pre-trained detector.
CIT is inspired by~\cite{Hall2014}. It is similar to CFT, except that the
trackers are warm-started from the pre-trained category-level detector.

\noindent\textbf{Results.}
As shown in Table~\ref{tab:rescar}, ODAMOT outperforms all related methods
that rely on the same general-purpose detectors trained on Pascal VOC 2007.
As expected, unrelated training data strongly degrades MOT performance.
Nevertheless, our results show that domain adaptation partly mitigates this
problem. By improving recall and maintaining high precision, ODAMOT allows
to track more targets than the related CFT and CIT online methods, which do
not perform the \emph{joint} adaptation of category and instance models.
This multi-task online adaptation allows to gradually discover and track
more targets while limiting model drift, although at the cost of moderately
increased identity switches and track fragmentation. On the other hand,
off-line ABT-type methods (DP\_MCF~\cite{Pirsiavash2011} and
G\_TBD~\cite{Geiger2014}) suffer greatly from the low quality of the
pre-trained detector, especially when using "mini-FV" (upper block of
Table~\ref{tab:rescar}).
As expected, more powerful state-of-the-art ConvNet features improve all
results (from $19.7\%$ to $23.6\%$ MOTA for ODAMOT) but surprisingly not
substantially. This confirms the difficulty of domain transfer, in
particular due to the overfitting tendency of deep nets, which is
problematic when faced with dataset bias. Note that this might be partly
alleviated by using features from earlier layers, which might transfer
better~\cite{Yosinski2014}. Our results also hint at the transferability
potential of the weaker mini-FV features, where ODAMOT improves more
significantly the MOT performance \wrt the baselines.

\noindent\textbf{Failure cases.}
Our method tends to suffer from two main problems. The first is tied to the
failure modes of the detector (missed detections and false alarms), and is
common to all TBD methods. Although our adaptation improves, the multi-task
objective tends to favor conservative updates to prevent drift, similarly
to self-paced learning approaches like~\cite{Tang2012}.
Second, our tracks contain many ID switches and are generally fragmented.
This hints at a lack of specialization of the appearance models, which
could be addressed by designing features that can better represent instances.
Another solution to this issue would consist in complementing our method
with advanced inter-target and occlusion reasoning, \eg~\cite{Yoon2015}.

\section{Conclusion}

We address the question of how to re-use object detectors pre-trained on
general-purpose datasets for causal multi-object tracking, when strongly
related training data is not available. To overcome the dataset bias
present in these generic detectors, we propose the joint online adaptation
of category- and target-level detectors. 
Our multi-task adaptation from category-to-instances and back allows to
improve overall MOT accuracy by increasing recall while maintaining high
precision and limiting model drift in challenging real-world videos.

\bibliography{references}

\end{document}